\documentclass[letterpaper, 10 pt, conference]{ieeeconf}  
\IEEEoverridecommandlockouts                             
\usepackage{amsmath,amssymb,amsfonts}
\usepackage{algorithmic}
\usepackage{graphicx}
\usepackage{url}

\usepackage{booktabs}
\usepackage{siunitx}
\usepackage{lipsum}

\usepackage{algorithm}
\usepackage{algorithmic}
\usepackage[ruled,algo2e]{algorithm2e}
\SetEndCharOfAlgoLine{} 
\usepackage{textcomp}
\usepackage{float}
\usepackage{booktabs}
\let\labelindent\relax
\usepackage{enumitem} 
\usepackage{multirow}

\def\BibTeX{{\rm B\kern-.05em{\sc i\kern-.025em b}\kern-.08em
    T\kern-.1667em\lower.7ex\hbox{E}\kern-.125emX}}
\usepackage{xcolor}
\usepackage{subcaption}  
\usepackage{svg}

\newcommand{\sketch}{s}
\newcommand{\depthimage}{I}
\newcommand{\dronepath}{w}
\newcommand{\autoproj}{p}
\newcommand{\toj}{^{(j)}}

\begin{document}

\title{\LARGE \textbf{SketchPlan: Diffusion Based Drone Planning From Human Sketches}}

\author{
    Sixten Norelius$^{1}$, Aaron O. Feldman$^{2}$, Mac Schwager$^{2}$%
    \thanks{$^{1}$EECS, KTH Royal Institute of Technology, Stockholm 115 59, Sweden
        {\tt\small sixtenn@kth.se}}%
    \thanks{$^{2}$Department of Aeronautics and Astronautics,
Stanford University, Stanford, CA 94305, USA
        {\tt\small aofeldma@stanford.edu, schwager@stanford.edu}}%
}
\maketitle

\begin{abstract}
We propose SketchPlan, a diffusion-based planner that interprets 2D hand-drawn sketches over depth images to generate 3D flight paths for drone navigation. SketchPlan comprises two components: a SketchAdapter that learns to map the human sketches to projected 2D paths, and DiffPath, a diffusion model that infers 3D trajectories from 2D projections and a first person view depth image. Our model achieves zero-shot sim-to-real transfer, generating accurate and safe flight paths in previously unseen real-world environments.  To train the model, we build a synthetic dataset of 32k flight paths using a diverse set of photorealistic 3D Gaussian Splatting scenes. We automatically label the data by computing 2D projections of the 3D flight paths onto the camera plane, and use this to train the DiffPath diffusion model.  However, since real human 2D sketches differ significantly from ideal 2D projections, we additionally label 872 of the 3D flight paths with real human sketches and use this to train the SketchAdapter to infer the 2D projection from the human sketch.  We demonstrate SketchPlan’s effectiveness in both simulated and real-world experiments, and show through ablations that training on a mix of human labeled and auto-labeled data together with a modular design significantly boosts its capabilities to correctly interpret human intent and infer 3D paths. In real-world drone tests, SketchPlan achieved 100\% success in low/medium clutter and 40\% in unseen high-clutter environments, outperforming key ablations by 20–60\% in task completion. The code for SketchPlan is publicly available at \url{https://github.com/sixnor/SketchPlan}.
\end{abstract}

\section{Introduction}
Instructing robots can be difficult for humans. Typical solutions include either low level teleoperation or high-level task specifications tailored for specific domains. This leaves a gap for loose structure interaction in a modality natural to humans. Natural language instruction has been successfully deployed in a variety of contexts \cite{huang2023voxposer, splatnav, lynch2023interactive, tan2025mobile} but may become cumbersome to prompt as execution requirements become more nuanced. For example, consider trying to guide a robot towards a specific patch of a monochromatic floor. Because this task lacks features that are easy to articulate in natural language, it would be hard to accomplish. In contrast, communicating instructions via a hand-drawn sketch can offer fine-grained spatial intent while still remaining intuitive and quick to formulate for humans. Such approaches have found success in manipulation based tasks \cite{gu2023rt, sundaresan2024rt} with an overhead or oblique camera view with few occlusions between objects. 
However, in contexts such as Unmanned Aerial Vehicle (UAV) navigation, the camera is typically egocentric, yielding a less informative perspective and making sketches harder to interpret via feature matching \cite{sundaresan2024rt} or direct unprojection \cite{gu2023rt, mehta2025l2d2} of the sketch into 3D. To overcome this challenge, imitation learning is a promising approach, but unfortunately, no dataset of real-world flights with matching sketches, egocentric images, and trajectories to learn from exists. Conducting a sufficient number of flights for a diverse set of environments would pose an exhausting endeavor. 

\begin{figure}
    \centering
    \includegraphics[width=\linewidth]{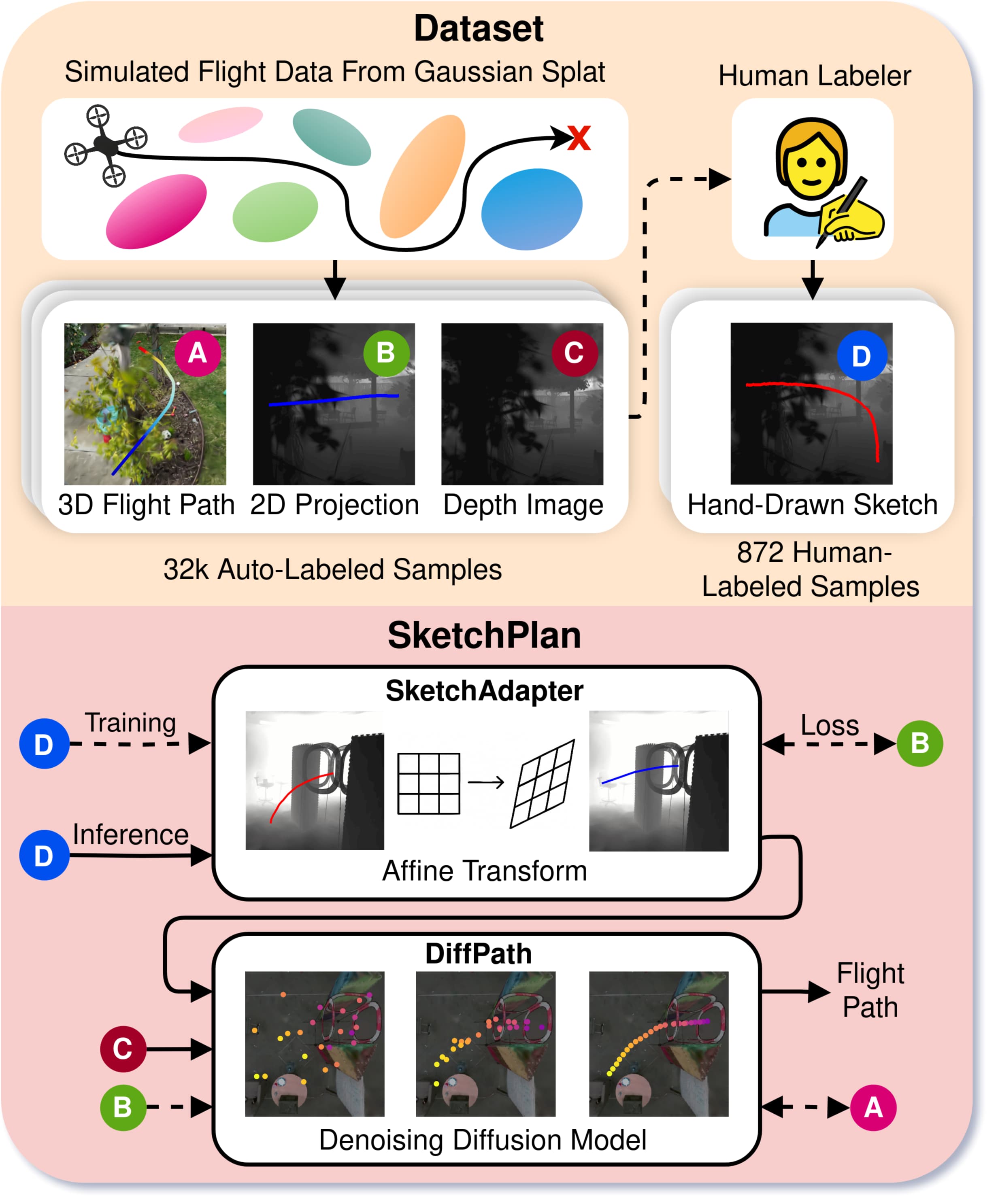}
    \caption{SketchPlan Architecture. SketchAdapter takes a 2D human sketch and predicts the intended 2D path projection, trained with a small set of human-labeled data.  DiffPath infers a 3D path from its 2D projection and an FPV depth image, trained with a large volume of auto-labeled 3D paths from diverse 3DGS scenes. Dashed arrows indicate loss computation for training.
    }
    \label{fig:systemdiag}
    \vspace*{-7mm}
\end{figure}
Motivated by this, we choose to forgo real-world flight data altogether and instead assemble a dataset composed entirely of simulated drone paths matched with depth images using 3D Gaussian Splatting (3DGS) \cite{3dgs} in a diverse set of environments. This allows for rapid generation of high-fidelity visual and spatial data, which we then label with hand-drawn sketches after the fact for a subset of the flights, creating a partially human-labeled dataset. 

Utilizing this dataset, we propose SketchPlan, a flexible human-sketch to 3D path planner. SketchPlan is trained by dividing the planner into two separate components, SketchAdapter and DiffPath, allowing for exploitation of drone paths both with and without matching human sketches by training each component on different subsets of the data. SketchPlan takes as input a depth image with a hand-drawn human sketch overlaid and outputs a drone path composed of a sequence of 3D waypoints. Lending from the scale and flexibility of our approach, SketchPlan achieves zero-shot transfer from simulated data to entirely unseen real-world environments. 

The main contributions of our approach are: 
\begin{itemize}
    \item We introduce SketchPlan, a learned planner that translates a human-drawn sketch, drawn on top of a depth image, to a 3D flight path in 0.4s, enabling live human-in-the-loop navigation.
    \item We train SketchPlan on a dataset of 32k 3D flight paths synthesized using photorealistic indoor and outdoor Gaussian Splatting scenes, where less than a thousand are human-labeled with matching sketches.
    \item We show that SketchPlan achieves zero-shot generalizability, solving real-world drone navigation tasks in cluttered, unseen environments on hardware.
\end{itemize}

\section{Related Work}
In this section, we highlight some prior work in sketch-based methods for robotic instruction, 3DGS for UAV navigation, diffusion-based robot planning, and learning from partially-labeled data.

 Sketching has been used in navigation tasks by drawing an approximate overhead map of the scene with a desired path for the robot to follow \cite{tan2025mobile, skubic2003sketch, xu2024robot}. To match features between the map and the real-world to enable navigation, in \cite{tan2025mobile}, this is done via use of pretrained Visual Language Models (VLMs), whereas in \cite{xu2024robot}, an occupancy map predictor is learned. Critically, the relation between the drawn sketch and path is inferred and not \textit{a priori} specified, allowing for more nuance in interpretation. While sketching a full map enables precise spatial reasoning in cluttered environments, its complexity precludes on-the-fly sketching as in our approach.

As in our work, other approaches have considered sketching a robot trajectory, particularly for manipulation tasks \cite{gu2023rt, sundaresan2024rt, yu2025sketch, mehta2025l2d2, zhi2024teaching}. Approaches like \cite{gu2023rt, sundaresan2024rt, zhi2024teaching} utilize hindsight trajectory sketching, where they annotate a pre-existing dataset with synthetic sketches in post, in order to train a policy that can then go from trajectory sketch to actions. In contrast, \cite{yu2025sketch, mehta2025l2d2} try to leverage sketching as a means for quicker collection of demonstration data for manipulation policies. Amongst these, \cite{gu2023rt, yu2025sketch, mehta2025l2d2} generate sketch training data by projecting 3D demonstration trajectories into 2D. However, projection may be a poor representation for how a human would draw sketches for the 3D demonstrations, particularly if given an uninformative camera perspective \cite{mehta2025l2d2} such as a first-person view in our case. This highlights two key challenges for our problem of sketch-based drone navigation. Due to the egocentric camera view, we cannot rely on projections as proxies for hand-drawn sketches. Thus, we must learn a general, image-dependent association between sketch and trajectory, necessitating a novel dataset annotated with human sketches. 

To collect this dataset, we rely on 3DGS \cite{3dgs} to generate photorealistic interpretable 3D representations of real-world scenes. In robotics, 3DGS has enabled large scale dataset generation \cite{mao2024dreamdrive, sousvide, chen2025grad}. In \cite{sousvide, chen2025grad}, 3DGS is leveraged as the scene representation for a drone dynamics simulator, which is then used to train policies that achieve low sim-to-real gap. 3DGS's photorealism also enables low sim-to-real gap for training UAV perception tasks like human recognition \cite{choi2025uavtwinneuraldigitaltwins}. The 3DGS representation has also been directly used for robot planning. To generate collision-free paths, \cite{splatnav, jin2024gsplannergaussiansplattingbasedplanningframework} incorporate the gaussian splats into trajectory optimization.

Using the 3DGS data, we train a diffusion model to infer 3D trajectories from their 2D projections on a camera plane, conditioned on a depth image. Diffusion models have recently been adapted for robotic planning and control \cite{chi2023diffusion, janner2022planning}, offering a powerful and flexible framework for modeling complex, multimodal trajectories. They can be trained via behavior cloning or reinforcement learning, and have shown success in tasks like visuomotor manipulation \cite{chi2023diffusion, wang2024equivariant, ze20243d} and goal-conditioned planning \cite{sridhar2024nomad, reuss2023goal}. Beyond training, inference-time guidance has been proposed to better align outputs with human intent \cite{wang2024inference}. Despite their strengths, diffusion models are prone to overfitting on small datasets \cite{he2025demystifying}, motivating the generation of a large auto-labeled dataset for greater generalization.

 We operate in a partially-labeled setting by using only a small subset of human-labeled data while training on a much larger auto-labeled dataset.  We train a lightweight  model to adapt from human-sketches to auto-labeled 2D projections. Recent works have applied similar strategies to enhance behavior cloning and skill transfer, such as mining positives from unlabeled data \cite{wang2023improving}, and generating synthetic labels using model confidence,  predictors, or image alignment \cite{zheng2023semi, sundaresan2024rt, park2022surfsemisupervisedrewardlearning, rozsypalek2022semi}.

\section{Dataset}
In this section, we describe how the training data is collected without requiring real-world examples of drone flights. This is enabled by exploiting 3D Gaussian Splatting (3DGS) models of several real-world environments. By using 3DGS environments, we can obtain accurate representations of both visual and geometric aspects of the scene. This accuracy results in minimal sim-to-real gap, enabling us to train using simulated data but then directly deploy our sketch-conditioned planner for real-world testing.

Each labeled sample of the dataset consists of a path $\dronepath^{(j)} \in \mathbf{R}^{T \times 3}$ represented by a sequence of 3D waypoints in the drone's body frame, a depth image $\depthimage\toj$, a 2D projection of the 3D path onto the camera plane $\autoproj^{(j)} \in \mathbf{R}^{T \times 2}$ and for a subset of samples, a human sketch $s^{(j)} \in \mathbf{R}^{T \times 2}$ drawn on top of $\depthimage\toj$. A relatively small subset of the data is human-labeled, i.e., has associated human sketches, but we also use the remaining large number of 3D path samples auto-labeled with 2D projections in our modular training approach.

\subsection{Environments}
We use the Zip-NeRF \cite{zipnerf} dataset, which contains high-quality images with registered poses in diverse environments, to train our model. Specifically, we use environments \texttt{alameda}, \texttt{berlin}, \texttt{london} and \texttt{nyc}, which feature a mix of indoor and outdoor scenes. Because the poses from these environments had different non-metric scalings, they were scaled to a shared metric frame by using common household objects with known measurements as reference (e.g., a video game console). This is crucial because the planner returns 3D paths in a metric frame. From the image and pose pairs, we produce 3DGS representations of the scenes using \cite{nerfstudio}. Although the raw images in the dataset contain no depth information, the 3DGS novel view synthesis capability enables the rendering of photo-realistic depth images from arbitrary poses of the drone camera. 
Moreover, the collision geometry obtained from 3DGS is high-quality, so that 3D paths which are collision-free with respect to the 3DGS will also be so in the ground-truth scene. We also reserve a test 3DGS environment (\texttt{flight}) similar to the real environments we use in hardware testing to assess how SketchPlan generalizes to an unseen environment.

\subsection{Path and View Generation}
Given a 3DGS of an environment, we wish to automatically generate a large dataset of collision-free 3D trajectories in this environment to then label with sketches. To do so, we sample collision-free start and end points densely within the gaussian splat. To find a feasible trajectory between the two points, we employ Splatnav \cite{splatnav}, a motion planning framework that generates collision-free, spline-parameterized 3D trajectories using a 3DGS environment representation. For data regularity, trajectories are clipped to a fixed arc length of 3 meters. A sequence of $T=100$ discrete waypoints $\dronepath^{(j)}$ is obtained by selecting uniformly in arc length within the clipped trajectory. In addition, the depth image $\depthimage^{(j)}$ is rendered by orienting with the trajectory start i.e., the camera is placed at the first 3D waypoint while facing the second and has the world vertical axis aligned with the image vertical. We also collect the 2D projections $\autoproj^{(j)}$ of the paths $\dronepath^{(j)}$ onto the camera plane (for details see \cite{hartley2003multiple}). 

In general, we find that our selection of environments yields good path diversity. The indoor scenes \texttt{nyc} and $\texttt{berlin}$ feature sharp turns around corners with poor visibility, while the outdoor scenes \texttt{london} and \texttt{alameda} have less abrupt turns and more weakly structured geometry such as plants, bushes and trees. 

\begin{table}[h]
    \centering
    \caption{Dataset metrics for the number of 3D paths and human sketches in each environment. \\ *: Extrapolated from 100 sketches. }
    \label{tab:dataset_metrics}
    \begin{tabular}{lcccc}
        \toprule
        Env. & \#Paths & \#Sketches & Plan Time & Sketch Time* \\
        \midrule
        \texttt{alameda} & 9711 & 242 & 50 min & 40 min \\
        \texttt{berlin} & 7488 & 203 & 47 min & 30 min\\
        \texttt{london} & 8813 & 223 & 48 min& 33 min \\
        \texttt{nyc} & 6550 & 204 & 37 min & 26 min \\
        \textbf{Total} & \textbf{32562} & \textbf{872} & \textbf{3 h 2 min} & \textbf{2 h 9 min} \\
        \texttt{flight} & -- & 151 &  --& -- \\
        \bottomrule
    \end{tabular}
\end{table}

\subsection{Sketching}
\begin{figure}[htbp]
    \centering
    \includegraphics[width=\linewidth]{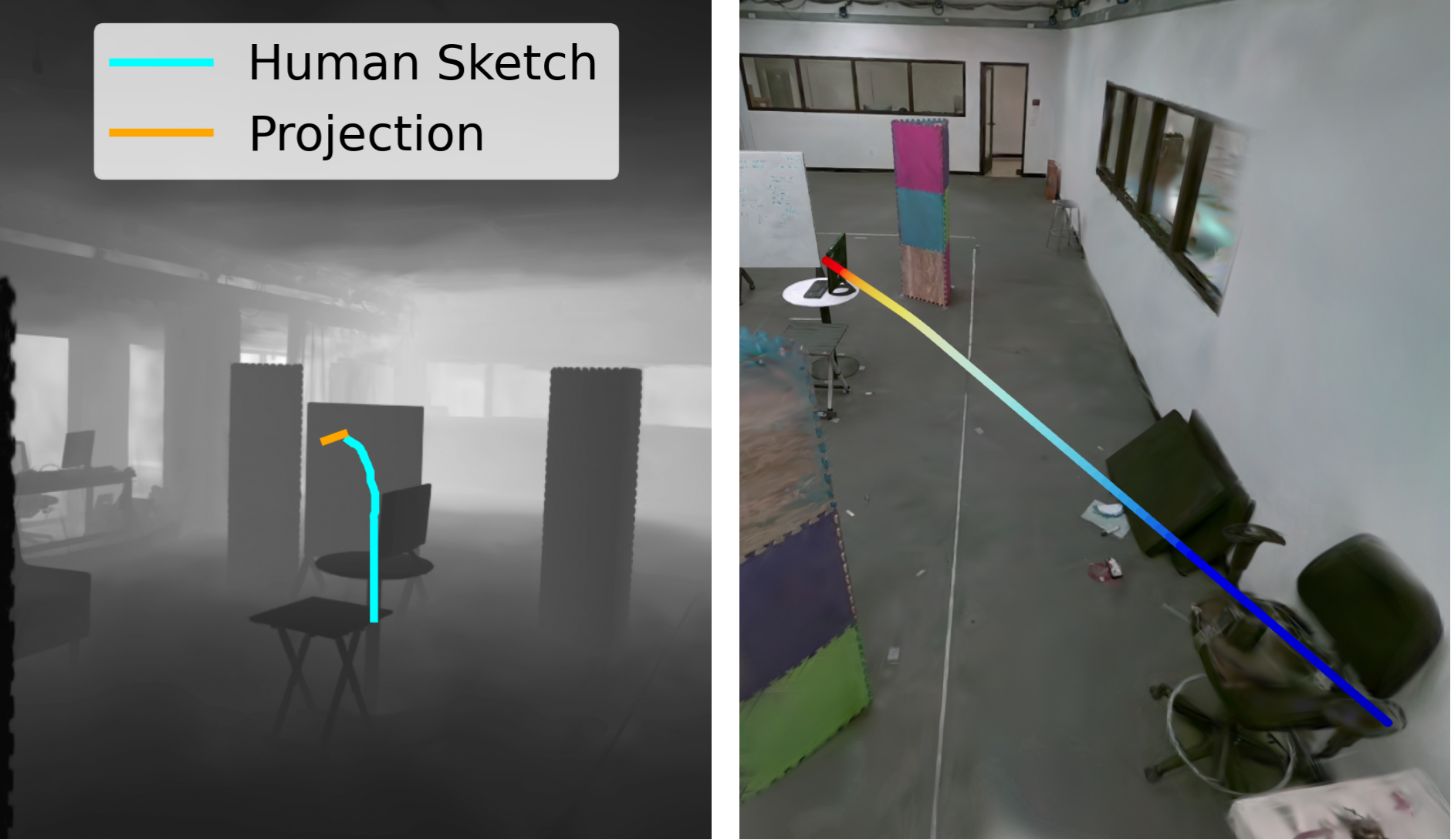}
    \caption{Comparison of human sketch and projection (left) for a 3D path (right). Note that the camera projection of the 3D path differs significantly from the human sketch. The projection maps the straight part of the path to a tight pixel cluster in the middle of the image while the human instead sketches a straight vertical line.}
    \label{fig:proj_vs_human}
    \vspace*{-5mm}
\end{figure}

Intuitively, it might be expected that for a given path $\dronepath^{(j)}$ and depth image $\depthimage^{(j)}$, the projection of the path $\autoproj^{(j)}$ and a corresponding human sketch $\sketch^{(j)}$ should be similar enough that collecting $\sketch\toj$ would be redundant. However, because of our egocentric camera viewpoint, $\autoproj\toj$ and $\sketch\toj$ are often very different. For instance, as shown in Fig. \ref{fig:proj_vs_human} an almost entirely straight 3D path gets projected to nearly a single point in the middle of the image. Thus for the dataset to accurately represent human intent in going from $\sketch\toj$ to $\dronepath\toj$, directly using $\autoproj^{(j)}$ as a proxy for $\sketch\toj$ does not suffice. 

Therefore, after producing the dataset of 3D trajectories in the various environments, we obtain corresponding 2D sketches by querying a human user/labeler. Given a sequence of waypoints $\dronepath^{(j)}$ and depth image $\depthimage^{(j)}$, the human draws a sketch matching $\dronepath^{(j)}$. For user convenience, $\dronepath^{(j)}$ was displayed in a 3D interactive viewer of the 3DGS environment, allowing the sketcher to accurately gauge how $\dronepath^{(j)}$ interacts with the scene geometry. The user simply draws on top of $\depthimage^{(j)}$ using a computer mouse or stylus to produce $\sketch^{(j)}$ (by resampling their drawing to a fixed number of points). 
At first, our approach to data generation may seem counterintuitive; although we wish to predict from input sketch $\sketch\toj$ the corresponding path $\dronepath\toj$, the human labels in reverse, taking input $\dronepath\toj$ and producing $\sketch\toj$. However, first sampling a 3D trajectory and then drawing the corresponding 2D sketch allows for quicker labeling and the generation of auto-labeled data (paths with projections but without sketches) with no human interaction. Thus, we can generate a large and diverse auto-labeled dataset of path and depth image pairs for model training. Since labeling all paths is time-prohibitive, we use a modular approach to train. All 32k paths are used in training, while the human only sketches labels for a small subset, taking just 2 hours, as shown in Table \ref{tab:dataset_metrics}. However, a resulting limitation of this approach is that the human may desire paths never produced in automatic path generation i.e., there is mismatch between the types of paths humans would like to execute and ones that can arise from the data generation method. We counteract this pitfall by using a flexible path planner and choosing diverse environments to promote path diversity.

\section{SketchPlan}
The sketch-to-path model, SketchPlan, is composed of two separate, independently trained modules. An overview of the architecture is presented in Fig. \ref{fig:systemdiag}. The first, SketchAdapter, ingests human sketches and infers the 2D camera projection of the 3D path. The second, DiffPath, aims to then reconstruct the 3D paths from the 2D camera projections. Since 2D projections are available for both the labeled and auto-labeled data, we train DiffPath on the full dataset. In contrast, SketchAdapter is trained only with the human labeled data. Thus, taken together SketchAdapter and DiffPath use both labeled and auto-labeled data in a partially-labeled learning approach. At inference, these components are chained together to go from human sketch to 3D path. Lastly, the paths generated by DiffPath tend to avoid collisions, but are not guaranteed to do so.  We therefore include a reactive collision avoidance controller on the drone as a safety filter on the paths generated by SketchPlan. 

\subsection{SketchAdapter}

The human sketch $\sketch\toj$ and projection $\autoproj\toj$ are often not very similar. However, we still expect them to be highly correlated. To bridge the gap between sketch and projection, we train our adaptation module, SketchAdapter, to go from $\sketch\toj$ to $\autoproj\toj$, outputting predictions $\hat\autoproj\toj$ on the human-labeled portion of the dataset. To avoid overfitting to the limited amount of labeled data, we opt to learn an affine transform between the two using Partial Least Squares (PLS) \cite{abdi2010partial}, a method which maximizes covariance between the input and output features in a latent space with a fixed number of components. We choose the number of components to be 5, resulting in a low-dimensional yet surprisingly expressive latent for the SketchAdapter model.

\subsection{DiffPath}
Our path generation model, DiffPath, ingests at inference the depth image $\depthimage\toj$ and approximate projection $\hat{\autoproj}\toj$ produced from SketchAdapter and returns a suitable 3D path $\hat{\dronepath}\toj$. There exist many suitable 3D paths for a particular 2D human sketch and image due to ill-posedness of the problem. Rather than explicitly/structurally biasing the model toward a class of suitable paths, DiffPath learns the distribution of 3D paths from training examples. The model can thus learn human-sketching biases through data. We use a diffusion model as the DiffPath backbone to allow for expressive path distribution modeling. 

\begin{figure}
    \centering
    \includegraphics[width=\linewidth]{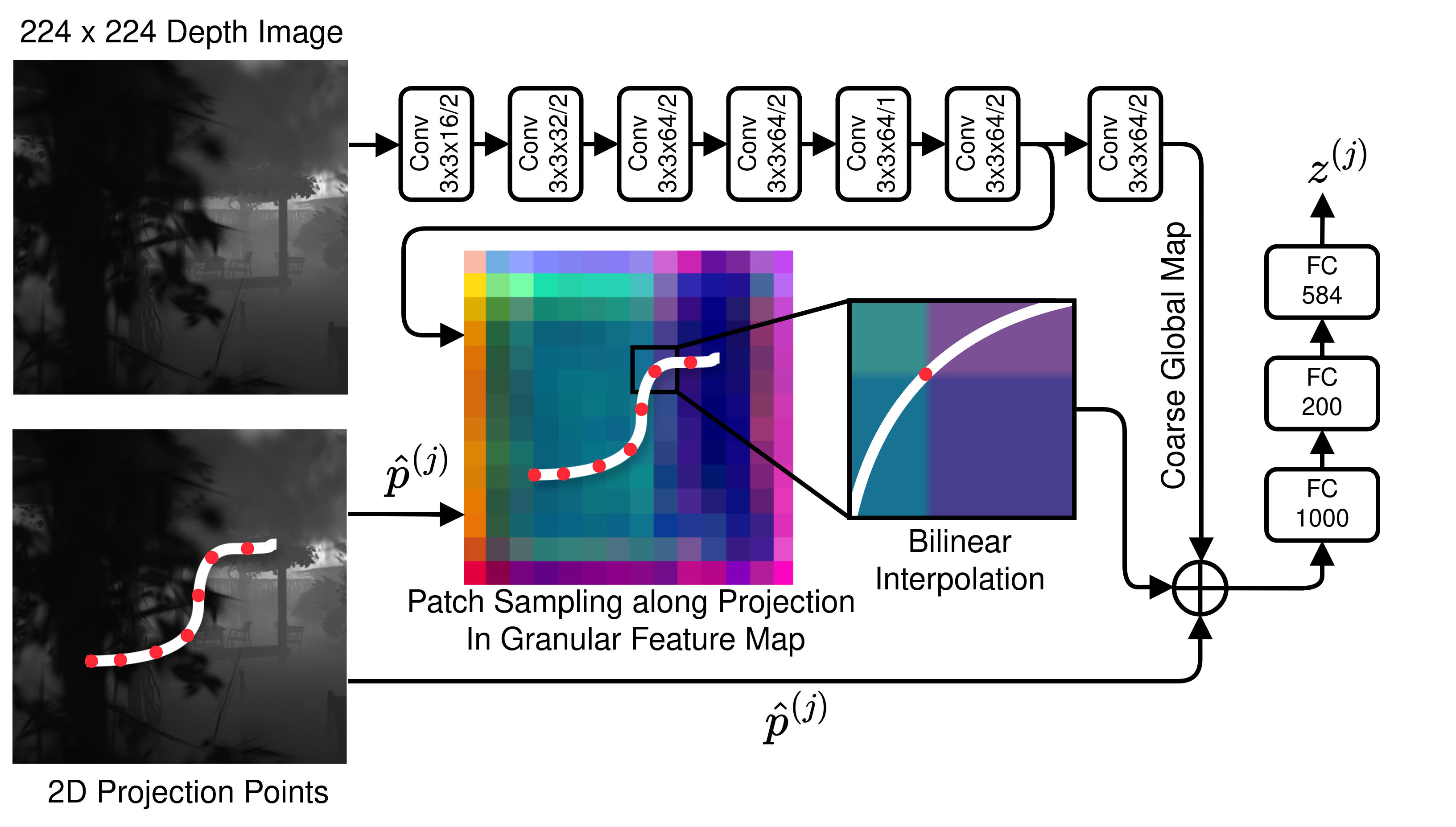}
    \caption{DiffPath Encoder. Our encoder combines the depth image and 2D projection into a compact embedding $z\toj$ by sampling patches along the projection on a granular feature map. Convolutional kernels denoted as $K\times K \times C /S$, where $K$ is the kernel size, $C$ the channels and $S$ the stride.}
    \label{fig:encoder}
    \vspace*{-5mm}
\end{figure}

\noindent\textbf{Encoder.}
Our vision encoder is based on a convolutional neural network (CNN). We rescale the depth images to size $224 \times 224$, normalize the pixel values to be zero mean with unit variance and clip the maximum depth at 15 m. We first pass the depth image $I\toj$ through 6 convolutional layers to produce an intermediate feature map. Then, we sample from the produced feature map along $\autoproj\toj$ (which has pixel coordinates) using bilinear interpolation, to obtain a set of the local feature embeddings. Notably, these local feature embeddings serve as intermediates that simultaneously capture information from the projection $\hat{\autoproj}\toj$ and the depth image. Finally, an additional convolutional layer is applied to the full feature map to obtain a coarse global embedding of the image. After this, the local, global and $\hat{\autoproj}\toj$ features are flattened and concatenated, to serve as inputs for an MLP that produces a final combined embedding $z\toj$ of $\hat{\autoproj}\toj$ and $\depthimage \toj$.

\noindent\textbf{Denoising Network.}    
We utilize a 1D UNet with Feature-wise Linear Modulation (FiLM) \cite{perez2018film} conditioning as the denoising network as in \cite{chi2023diffusion}. To condition the denoising network, we use the embedding $z\toj$ produced by our encoder. Thus, the diffusion denoising process depends on the provided projection and depth image.
We utilize a 1D UNet with Feature-wise Linear Modulation (FiLM) \cite{perez2018film} conditioning, as in \cite{chi2023diffusion}, to model the denoising process. At inference time, DiffPath starts from a purely noisy path sample $\hat\dronepath_T\toj \sim \mathcal{N}(0, I)$ and by conditioning on $z\toj$ iteratively refines it toward a coherent 3D trajectory through a sequence of denoising steps:

\begin{equation}
\label{eq: denoising}
    \hat\dronepath_{t-1}\toj = F_\theta\left(\hat\dronepath_{t}\toj,t,z\toj\right)
\end{equation}
where $F_\theta$ is the stochastic denoising function parametrized by $\theta$ and $t$ is the denoising step. Each step incrementally shapes the path to be consistent with the scene and user intent encoded in $z\toj$, eventually reaching a denoised predicted path $\hat\dronepath_{0}\toj=\hat\dronepath\toj$. 

During training, we jointly optimize both encoder and denoising networks. The latent representation $z\toj$ is constructed using the auto-labeled 2D projection $\autoproj\toj$ (instead of $\hat{\autoproj}\toj$). The denoising network is trained to reconstruct the original 3D path $\dronepath\toj$ in the dataset from a perturbed version by minimizing mean squared error (see \cite{chi2023diffusion} for details).

\subsection{Collision Filter}
Although DiffPath conditions on the depth image and usually avoids collisions, the generated path waypoints are not guaranteed to be collision free. To remedy this issue at test time, we postprocess the generated waypoints with a simple collision filter. If a generated waypoint $\hat{w}_i$ is in collision with the depth image, the filter finds a modified collision-free waypoint $\tilde{w}_i$ close to $\hat{w}_i$. To identify collision, we use the depth images to create a live 3D point cloud at runtime. More specifically, we retain the last 10\,s of depth images, unproject the pixels in each depth image into a 3D point cloud with points $\{p_i\}_{i=1}^N$ and combine and filter for outliers. To tractably find a collision-free $\tilde{w}_i$, we first find a collision-free convex polytope around the current robot position $x$ in which to search. 

\begin{algorithm2e}[ht]
\caption{Safe Polytope Construction}
\label{alg:safepolytope}
\KwIn{Current position $x$, point cloud $\{p_i\}_{i=1}^{N}$, robot radius $r$}
\KwOut{Set of halfspace constraints $\mathcal{H}$}

$\mathcal{P} \gets \{p_i\}_{i=1}^{N}$\tcp*[r]{Init unchecked points}
$\mathcal{H} \gets \varnothing$\;

\While{$\mathcal{P} \neq \varnothing$}{
    $p_{\mathrm{close}} \gets \mathrm{NearestNeighbor}_{\mathcal{P}}(x)$\;
    $a \gets (p_{\mathrm{close}} - x)/\Vert p_{\mathrm{close}} - x \Vert$\;
    $b \gets a^\top p_{\mathrm{close}} - r$\;
    $\mathcal{H} \gets \mathcal{H} \cup \{(a, b)\}$\;
    $\mathcal{P} \gets \{p \in \mathcal{P} \mid a^\top p \leq a^\top p_{\mathrm{close}}\}$\tcp*[r]{Prune}
}
\end{algorithm2e}
\vspace*{-4mm}

For each obstacle point \( p_i \), we define a supporting hyperplane orthogonal to the vector \( p_i - x\) and tangent to the \( r \)-ball centered at \(p_i\).
A waypoint \( \tilde{w}_i \) is guaranteed to be collision-free if it lies outside all such half-spaces, but some can be pruned while remaining safe, as we detail in Algorithm \ref{alg:safepolytope}. 

\begin{equation}
\begin{aligned}
    \min_{\tilde{w}_i} \quad & \Vert \tilde{w}_i - \hat{w}_i \Vert^2_2 + \alpha\Vert t \Vert_1 \\
    \text{subject to} \quad & a^T_i\tilde{w}_i \leq b_i+t_i, \quad (a_i,b_i) \in \mathcal{H} \\
                            & t_i \geq 0
\end{aligned}
\label{eq:collision_filter}
\end{equation}

Once this collision-free polytope has been formed, we solve for $\tilde{w}_i$ by minimizing the distance to $\hat{w}_i$, while adhering to the half-space constraints but with slack $t$ to obtain a solution if this is infeasible. This yields a quadratic program that can be solved at least 50 Hz on the drone, shown in (\ref{eq:collision_filter}).

\section{Simulated Experiments}
We test SketchPlan on the unseen simulation environment  \texttt{flight} to see how well it performs on a novel 3DGS scene, like those in the training dataset.

\subsection{Setup}

 The test setup in simulation is composed of an interactive sketching interface, where novel views are rendered from a 3DGS representation of \texttt{flight} upon which the user can draw a sketch to generate a path using SketchPlan. The collision filter is disabled in order to gauge how well the model implicitly plans collision-free paths. For evaluation, we utilize both new sketches without ground-truth 3D paths and 3D paths labeled with sketches, to gather a mix of qualitative and quantitative results.

\subsection{Metrics}
For the simulated testing, we define two quantitative metrics. Path Distance (PD), measures how much the predicted path $\hat{w}\toj$ differs from the ground truth $\dronepath\toj$, defined as 
\begin{equation}
    \mathrm{PD} = \sum_{i=1}^{T} \Vert \hat{w}\toj_i-w_i\toj \Vert_2.
\end{equation}

To gauge how well our model implicitly achieves collision avoidance, we check whether the robot would have collided with the 3DGS environment (leveraging the geometric interpretation of 3DGS as a union of ellipsoids as in \cite{splatnav}). We define the Collision Rate (CR) as the fraction of paths that collide with the environment. 

\subsection{Ablations}

To gauge the contribution of each component in our pipeline, we conduct a series of ablations to isolate the effect of individual design choices on overall performance.

\noindent\textbf{w MLP SketchAdapter.} The PLS SketchAdapter is replaced by an MLP with two hidden layers of dimension 256 each trained with BatchNorm and a dropout rate of $0.7$. 

\noindent\textbf{w/o Diffusion Model.} This ablation consists of the encoder directly predicting $\hat{\dronepath}\toj$ without passing through the diffusion model. This is accomplished by adding an additional affine layer for regression after the encoder has produced the embedding $z\toj$ of the input features. 

\noindent\textbf{w/o SketchAdapter.} The PLS SketchAdapter is removed entirely and the human sketch $\sketch\toj$ is fed directly to DiffPath and treated as if it were the projection $\autoproj\toj$. 

\noindent\textbf{w/o Auto-labeled data.} We train DiffPath on the raw human sketches from the labeled portion of the dataset, forgoing SketchAdapter and auto-labeled 2D projection data entirely. 

\noindent\textbf{Inference-Time Stochastic Sampling.} To test whether it is valuable to condition the diffusion explicitly on the predicted projection $\hat\autoproj\toj$, we test only conditioning on the depth image and instead utilize cost-guided Stochastic Sampling (SS) diffusion \cite{du2023reduce, wang2024inference}. At test time, we add an additional term to the denoising process (Eq.~\ref{eq: denoising}) to guide the generated 3D path $\hat\dronepath\toj$ to align with $\hat\autoproj\toj$ i.e., to minimize the MSE between the 2D projection of $\hat\dronepath\toj$ and $\hat\autoproj\toj$. 
\subsection{Results}

\begin{figure}
    \centering
    \includegraphics[width=\linewidth]{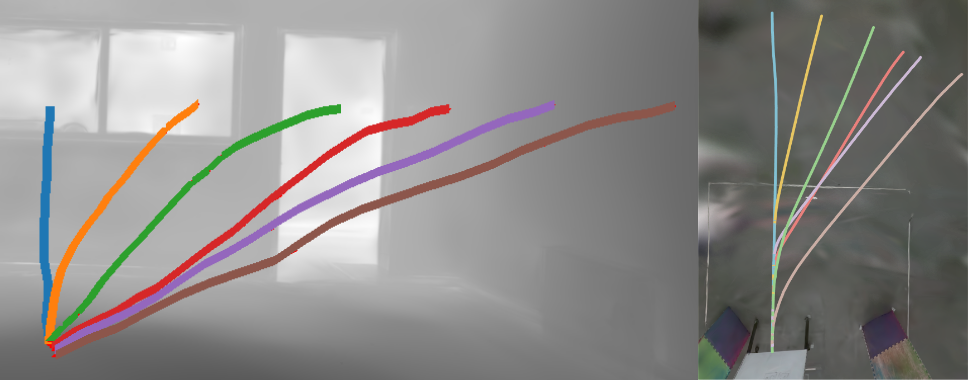}
    \caption{Left: Sample Sketches -- Right: Resulting 3D Paths. Note that more curvature in the sketch results in greater side translation in the generated 3D path. }
    \label{fig:senstest}
    \vspace*{-3mm}
\end{figure}

\begin{figure}
    \centering
    \includegraphics[width=\linewidth]{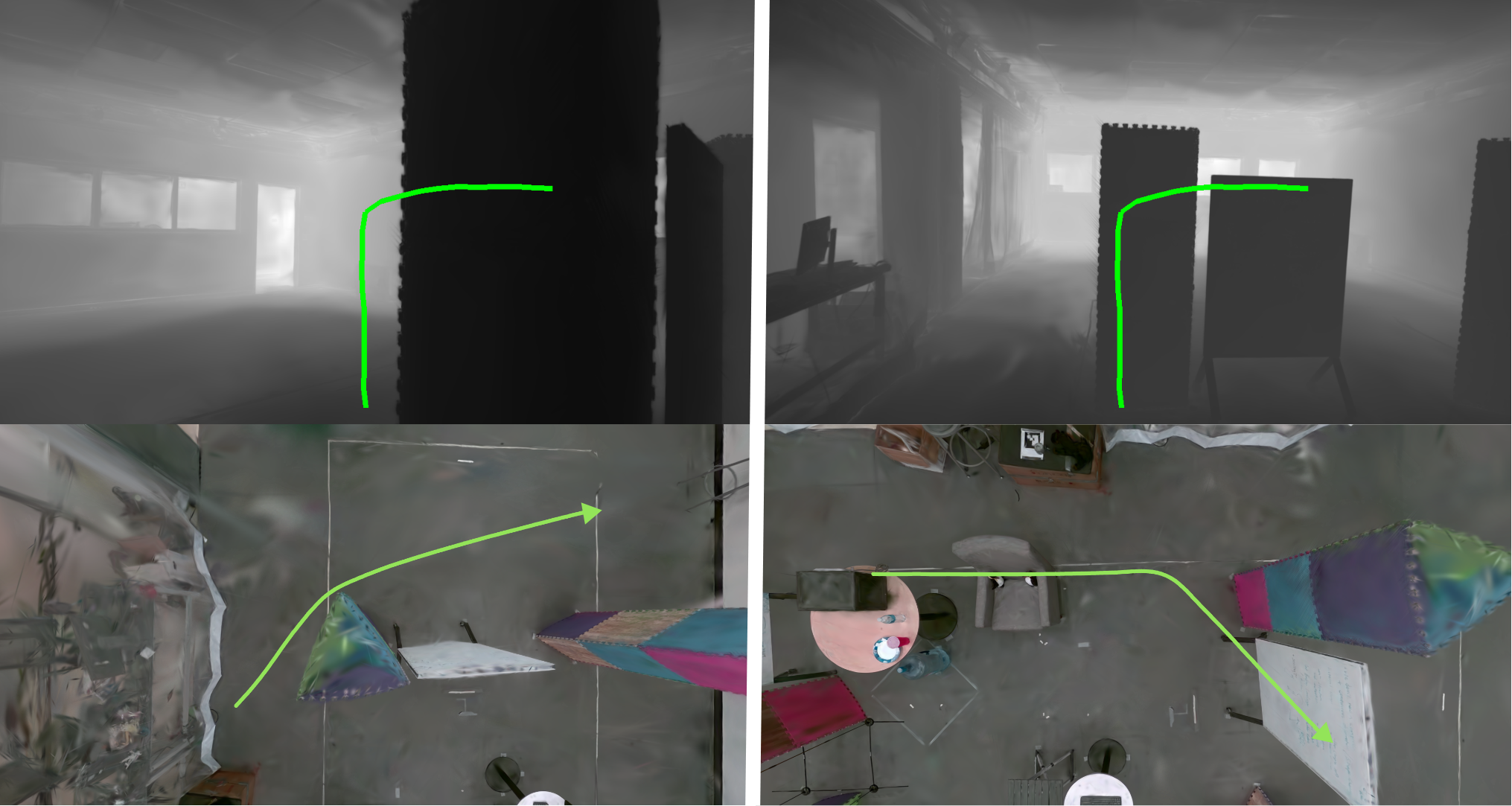}
    \caption{Rollout of the same 2D sketch in two different poses. Note how the curvature of the 3D path (bottom) changes depending on the background scene (top). Arrows in bottom panels denote path direction.}
    \label{fig:obstaclechangetest}
    \vspace*{-6mm}
\end{figure}

We find that SketchPlan takes on average 0.4s to generate a path in simulation on an NVIDIA RTX 4090, allowing it to be used in real-time human-in-the-loop operation.

In Figures \ref{fig:senstest} and  \ref{fig:obstaclechangetest}, we qualitatively test how SketchPlan adapts to changes in both environment and sketch. In Fig.~\ref{fig:senstest} while keeping the depth image fixed, we varied the drawn sketch's curvature. Observe that as the curvature of the sketch increases, so does that of the path, reflecting responsiveness to the human sketch. However, SketchPlan can still struggle with sketches where the intent is to generate paths with sharp turns or complex curvature, as these are rarely generated by SplatNav \cite{splatnav} and thus are rare in our dataset. This can can particularly be seen in Fig. \ref{fig:obstaclechangetest}, where the sketch intent was to generate an almost $90$\textdegree-turn, which SketchPlan fails to do. 

Meanwhile, keeping the sketch fixed as in Fig. \ref{fig:obstaclechangetest} but changing the environment, and the associated depth image presented to SketchPlan, also yields different paths. Specifically, we observe that the generated 3D path curvature changes based on the obstacle locations. This finding showcases SketchPlan's capabilities to use the environment to infer more granular human intent and implicitly avoid collision. Together, the qualitative results from Figures~\ref{fig:senstest} and \ref{fig:obstaclechangetest} showcase that our model learns behavior from both the environment and sketch, and that it would not suffice to use just one of them to predict the path.

\begin{table}[h]
    \centering
    \caption{Quantitative metrics for \texttt{flight} environment}
    \label{tab:ablation}
    \begin{tabular}{lcccc}
        \toprule
        Method & PD & Collision Rate \\
        \midrule
        Ours & 6.38 & 10.7\% \\
        w/o SketchAdapter & 17.4 & 27.8\% \\
        w MLP SketchAdapter & 6.18 & 12.9\% \\
        w/o Auto-labeled data & 10.99 & 17.4\% \\
        w/o Diffusion Model & 6.93 & 16.2\% \\
        Inference-Time SS \cite{wang2024inference} & 10.81 & 14.3\% \\
        \bottomrule
    \end{tabular}
    \vspace*{-4mm}
\end{table}

We also quantitatively assessed our method and compared against several ablations on withheld human-labeled validation data.

 From Table \ref{tab:ablation}, we note that SketchAdapter is critical as without it (\textit{w/o SketchAdapter)} both PD and CR are significantly higher, confirming that the human sketch and projections are indeed different with an egocentric camera view. The exact form of SketchAdapter seems to be less important, as we achieve similar metrics when using a shallow MLP (\textit{w MLP SketchAdapter)} instead of PLS. We expect that as the number of sketches increases, deeper models for SketchAdapter could perform better.

Training without auto-labeled data (\textit{w/o Auto-labeled data}) also degrades both PD and CR, which we suspect is due to the diffusion model overfitting.
In addition, we note that removing the diffusion model and directly using the encoder (\textit{w/o Diffusion Model}) to predict the path yields a higher CR. 

Finally, we find that using guidance instead of conditioning the model on the sketch (\textit{Inference-Time SS}) does not fail catastrophically but lags behind our approach in both PD and CR. This finding suggests that training DiffPath with 2D projection data is beneficial, instead of only using the inferred 2D projection $\hat\autoproj\toj$ at test time.

\begin{figure*}[http]
    \centering
    \includegraphics[width=\linewidth]{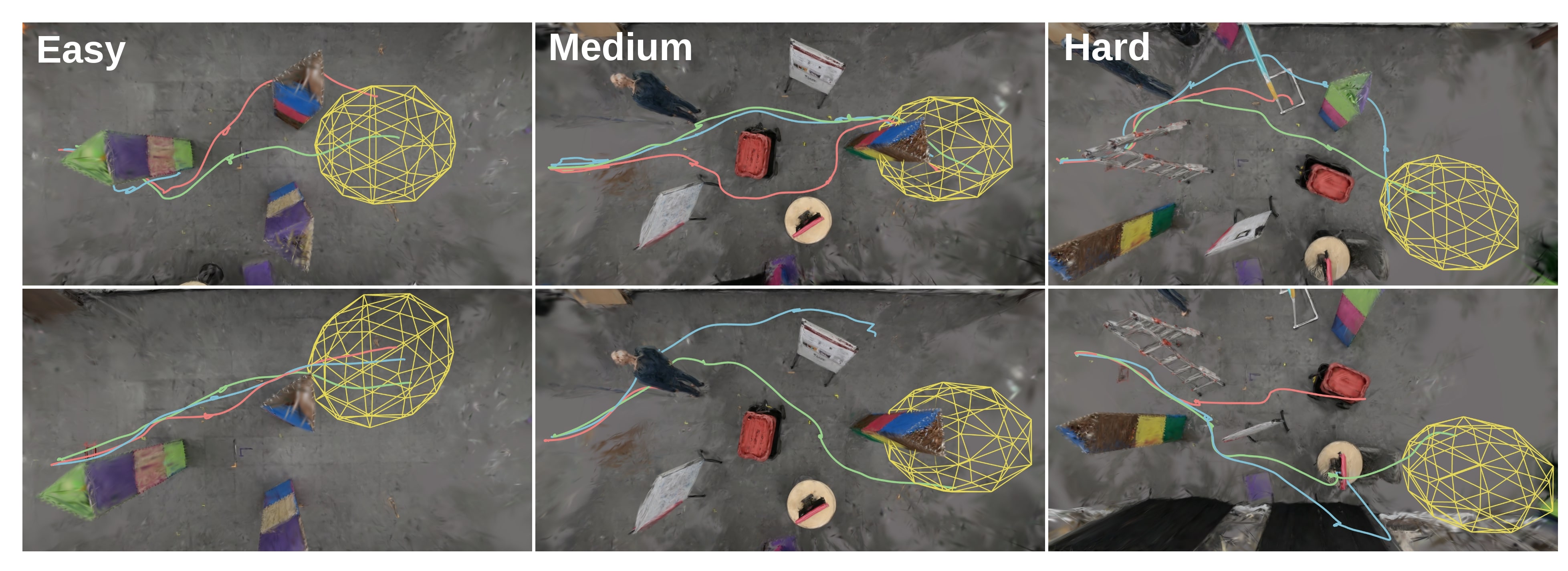}
    \caption{Sample drone hardware trials for the three environments. Green -- Ours, Blue -- w/o Diffusion, Red -- w/o SketchAdapter. 
    The wireframe ball indicates the goal region. }
    \label{fig:trials}
    \vspace*{-4mm}
\end{figure*}

\section{Drone Hardware Experiments}
We also evaluated our method on unseen real-world environments, demonstrating successful sketch-based navigation with a quadrotor drone. We trial on multi-sketch navigation tasks in three different environments of varying difficulty. Although our method is trained purely with 3DGS simulation, we find that it can be successfully applied zero-shot, overcoming the sim-to-real gap.

In each trial, the task involves navigating from a 3D start to a goal position, by only drawing sketches and having the drone track the corresponding waypoints generated by SketchPlan. To accomplish this, the user draws a sketch, SketchPlan is inferenced and the drone executes the path to then hover in place awaiting another sketch, repeating until the task is completed or failed. A trial is concluded if the goal is reached or the drone enters a state from which the user cannot recover using SketchPlan.
\subsection{Setup}
We construct three environments to reveal how our approach fares under different levels of obstacle clutter, height, and shape. Our \texttt{Easy} environment is composed of three triangular foam prisms arranged sparsely, acting as a test of SketchPlan's capabilities in a simple uncluttered scene. For \texttt{Medium}, more obstacles are added and arranged more densely. We also add obstacles with a lower height to test the capabilities of the planner to traverse overhead and not only to the side of obstacles. \texttt{Hard} features more intricate geometry such as a ladder and flight gate and is densely occupied. \texttt{Hard} was designed to be out of distribution from the training dataset, which generally contains sparsely occupied environments with simple geometry and limited verticality, to test the limits of SketchPlan. 

Our hardware setup consists of a quadrotor drone equipped with a Pixracer flight controller, Nvidia Jetson Orin Nano and ZED Mini stereo camera, from which we obtain a depth image of the current egocentric quadrotor view. All computation and model inference is off-board by transmitting the depth images over Wi-Fi, and sending waypoints back to the drone. Since our model requires connection to a host device for sketching, conducting inference off-board does not induce an additional need for computational resources.

\begin{table}[http]
\caption{Real-World Testing Metrics Across Difficulties and Models}
\label{tab:realres}
\centering
\begin{tabular}{llccccc}
\toprule
Env. & Model & Finish & Alt. & Avoid & Side & Sketches \\
\midrule
\multirow{3}{*}{\texttt{Easy}} 
    & Ours           & 5/5    & 4/5       &   17 cm    &    5/5   &    2    \\
    & w/o DP  & 2/5  & 1/2 &   11 cm    & 2/2  &   2     \\
    & w/o SA &3/5& 1/3 & 23 cm   &    2/3&   2.33     \\
\midrule
\multirow{3}{*}{\texttt{Medium}}
    & Ours &    5/5     &   4/5    &   30 cm    &      4/5 &  3      \\
    & w/o DP  &   2/5     &   2/2     &  39 cm    &  1/2  &  3      \\
    & w/o SA &     3/5    &  2/3     &   33 cm    &     1/3  &     3   \\
\midrule
\multirow{3}{*}{\texttt{Hard}}
    & Ours         & 2/5          &   2/2    &   29 cm    &   2/2    &  3      \\
    & w/o DP  &     1/5   &   1/1     &  36 cm    &    0/1 &  4      \\
    & w/o SA &   1/5      &    1/1   &   23 cm    &    1/1   &     3   \\
\bottomrule
\end{tabular}
\vspace*{-4mm}
\end{table}

\subsection{Metrics}

For each task attempt, the user is assigned a goal position to reach using sketch-controlled navigation. They then interact with the drone using continuous sketching to draw a sequence of sketches to attempt to reach the goal.

At the most basic level, we measure whether the task was completed at all and how many sketches it took using continuous sketching. We define \textbf{Finish} to be when the drone enters within 1 m of the goal location, which is a predetermined point in 3D space, visualized by wireframe balls in Fig. \ref{fig:trials}. Then, \textbf{Sketches} is the number of sketches it took in total to go from start to goal.

For the trials where the task was successfully completed, we record additional metrics related to whether human intent was mirrored in the executed paths, both in terms of topology and distance deviation. After the user had drawn a sketch but before the drone had started executing the predicted path, the human noted their intent according to the follow criteria:
\begin{description}[style=nextline, labelwidth=!, leftmargin=2.1cm, widest=Altitude]
    \item[\textbf{Side}\label{metr:side}] 
    On which sides should obstacles be passed? \\
    \textit{right / left / above / below}

    \item[\textbf{Altitude}\label{metr:altitude}] 
    What is the intended altitude? \\
    \textit{ascend / level / descend}
\end{description}
    Following path generation and tracking for all sketches in a run, these metrics were then scored to see whether user intent was fulfilled. For \textbf{Altitude}, we define \textit{ascend} (\textit{descend}) to be whether the drone's final position has increased (\textit{decreased}) by 30 cm or more compared to its starting position.
    
    We also collect a metric \textbf{Avoid} to see whether collision avoidance noticeably modified the original path generated by DiffPath by calculating the maximum deviation between the predicted waypoints and the collision avoidance adjusted ones for each sketch.  

\subsection{Results}

The results of the trials are noted in Table \ref{tab:realres}. The metrics beyond task completion (\textbf{Altitude, Avoid, Side}) are only presented for the trials that finished successfully. The results indicate that our full method outperforms the baseline ablations. SketchPlan manages to complete all trials successfully in the \texttt{easy} and \texttt{medium} environments. While it successfully reaches the goal less often in the \texttt{hard} environment, it still manages to satisfy the \textbf{Altitude} and \textbf{Side} metrics in cases where it does succeed. The typical failure mode for our method in the \texttt{hard} environment was to get stuck in front of the ladder and flight gate. We hypothesize this issue is because these obstacles have small openings surrounded by thin edges, which don't occur in the training environments. In contrast, without the diffusion model, the drone tended to generate highly noisy paths and veer from the sketch intension. Without SketchAdapter, the quadrotor tended to go out of bounds, especially towards the ceiling. This indicates that it is misinterpreting the intent of the user, as also reflected in the \textbf{Side} and \textbf{Altitude} metrics.

\section{Conclusion}
We presented SketchPlan, a diffusion-based drone path planner capable of converting human sketches drawn on depth images into viable 3D trajectories. By leveraging 3D Gaussian Splatting environments, we sidestep collecting real-world flight data, enabling scalable and diverse synthetic training. Our two-module design of SketchAdapter and DiffPath enables mixed-data training, using both human-labeled and auto-labeled data, and robust generalization. Simulated and real-world evaluations confirm that SketchPlan captures nuanced human intent and adapts to environmental context, outperforming ablations in path accuracy and task success. Our results demonstrate the feasibility and advantages of using sketch-conditioned diffusion models for intuitive UAV navigation in complex settings.

\noindent\textbf{Limitations.} 
 Our method is trained on sketches drawn to match pre-generated paths, which may not reflect the full range of paths that humans could intend to generate. This limits the types of paths SketchPlan can generate, such as correctly interpreting sketches with complex obstacle interaction or curvature. Moreover, all sketches in our dataset are drawn by the authors, which may not be representative for a broader set of users. While our collision filter reactively helps ensure safety, it cannot replan globally, and thus may get stuck in situations from which the user cannot sketch to recover. Future work could address these concerns by 1) taking human intent into account when generating the dataset, 2) collecting a more diverse set of sketches and 3) implementing a more adaptive replanning solution.

\bibliographystyle{IEEEtran}
\bibliography{IEEEabrv,bongo}

\end{document}